\documentclass[journal]{IEEEtran}
\usepackage{amsmath,amsfonts}
\usepackage{algorithmic}
\usepackage{algorithm}
\usepackage{tabularx,booktabs}
\usepackage{array}
\usepackage[caption=false,font=normalsize,labelfont=sf,textfont=sf]{subfig}
\usepackage{textcomp}
\usepackage{stfloats}
\usepackage{url}
\usepackage{verbatim}
\usepackage{graphicx}
\usepackage{cite}

\usepackage{ulem}
\usepackage{multirow}
\usepackage{color}
\usepackage[table,xcdraw]{xcolor}
\usepackage{makecell}
\usepackage{float}
\usepackage{amsmath} 
\usepackage{booktabs}

\makeatletter
\let\NAT@parse\undefined
\makeatother
\usepackage{hyperref}  

\begin{document}
\title{LMQFormer: A Laplace-Prior-Guided Mask Query Transformer for Lightweight Snow Removal}
\author{Junhong Lin, Nanfeng Jiang, Zhentao Zhang, Weiling Chen,~\IEEEmembership{Member,~IEEE} and Tiesong Zhao,~\IEEEmembership{Senior Member,~IEEE}
\thanks{This work was supported in part by the National Natural Science Foundation of China (Grant No. 62171134) and in part by Natural Science Foundation of Fujian Province, China (Grants No. 2022J02015 and 2022J05117). (Corresponding author: Tiesong Zhao.)}
\thanks{J. Lin, N. Jiang, Z. Zhang and W. Chen are with the Fujian Key Lab for Intelligent Processing and Wireless Transmission of Media Information, College of Physics and Information Engineering, Fuzhou University, Fuzhou 350108, China (E-mails: jhlin$\underline{~}$study@163.com, jnfrock@gmail.com, 211120091@fzu.edu.cn, weiling.chen@fzu.edu.cn).}
\thanks{T. Zhao is with the Fujian Key Lab for Intelligent Processing and Wireless Transmission of Media Information, College of Physics and Information Engineering, Fuzhou University, Fuzhou 350108, China and also with the Peng Cheng Laboratory, Shenzhen 518055, China (e-mail: t.zhao@fzu.edu.cn).}
}

\maketitle

\begin{abstract}
Snow removal aims to locate snow areas and recover clean images without repairing traces. Unlike the regularity and semitransparency of rain, snow with various patterns and degradations seriously occludes the background. As a result, the state-of-the-art snow removal methods usually retains a large parameter size. In this paper, we propose a lightweight but high-efficient snow removal network called Laplace Mask Query Transformer (LMQFormer). Firstly, we present a Laplace-VQVAE to generate a coarse mask as prior knowledge of snow. Instead of using the mask in dataset, we aim at reducing both the information entropy of snow and the computational cost of recovery. Secondly, we design a Mask Query Transformer (MQFormer) to remove snow with the coarse mask, where we use two parallel encoders and a hybrid decoder to learn extensive snow features under lightweight requirements. Thirdly, we develop a Duplicated Mask Query Attention (DMQA) that converts the coarse mask into a specific number of queries, which constraint the attention areas of MQFormer with reduced parameters. Experimental results in popular datasets have demonstrated the efficiency of our proposed model, which achieves the state-of-the-art snow removal quality with significantly reduced parameters and the lowest running time. Codes and models are available at https://github.com/StephenLinn/LMQFormer.
\end{abstract}


\section{Introduction}

\IEEEPARstart{S}{now} seriously affects the visibility of scenes and objects. It usually leads to poor visual qualities and severe performance degradations in high-level computer vision tasks such as object detection and semantic understanding. However, it is difficult to capture unified patterns in snowy scenes due to their different patterns and transparency. Unlike other types of image noises \cite{chen2021robust,nie2022stereo,ma2022toward,zhang2022underwater}, snow seriously obscures the background and thus is difficult to be removed. How to recover clean images from snowy scenes is still a challenging issue.

\begin{figure}[ht]
    \centering
	\includegraphics[width=0.485\textwidth]{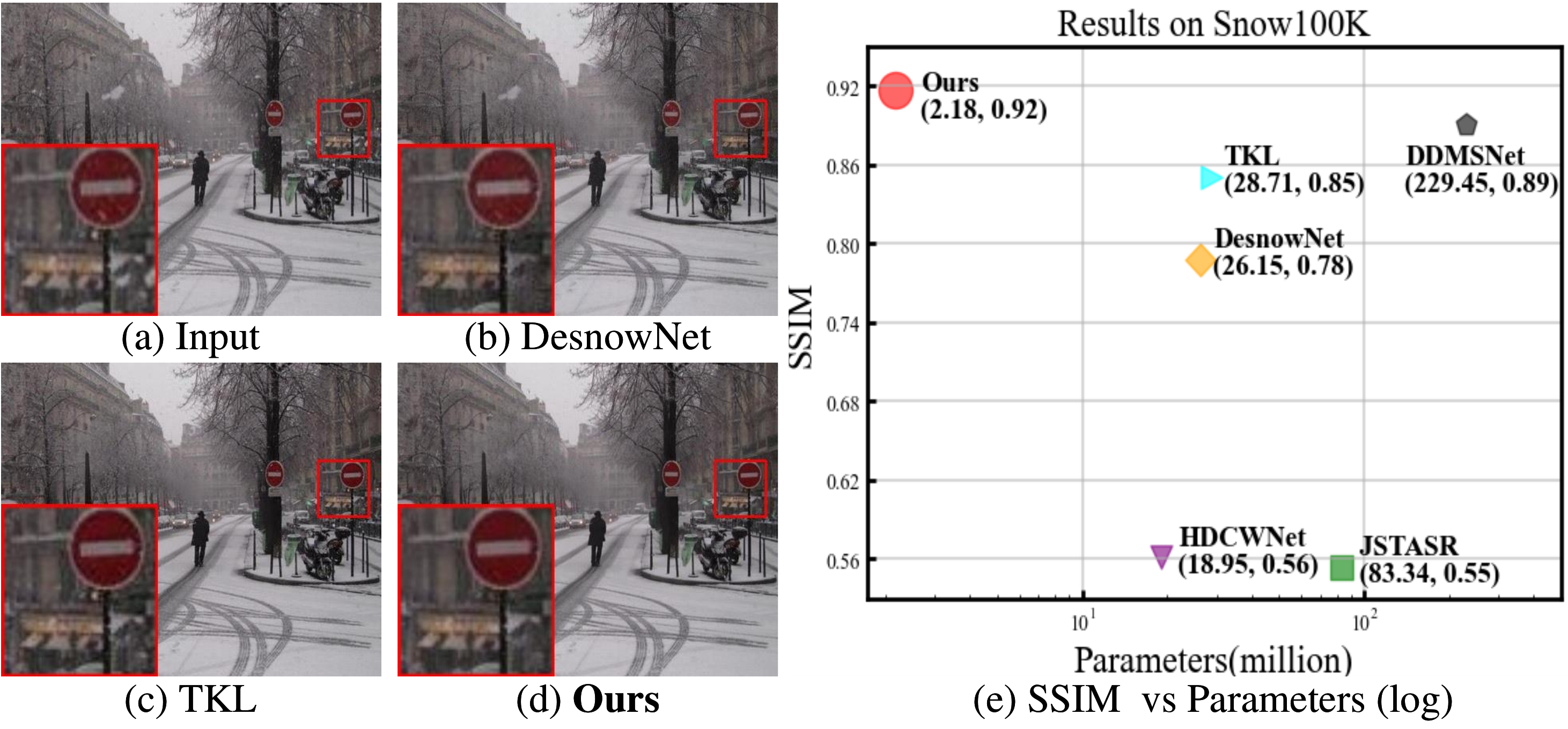}
	\caption{Our proposed method achieves the state-of-the-art snow removal quality with the lowest computational complexity. (a) A typical real-world snowy image. (b)-(d) the outputs of our method and its peers. (e) the average performances under Snow100K.}
	\label{Figure1}
\end{figure}

We divide existing snow removal methods into two types: traditional methods and deep-learning-based methods. Traditional methods are based on artificial prior knowledge to model snowy layers, such as HOG and MoG model \cite{2019Rain}, dictionary learning \cite{2017A}, color assumptions \cite{pei2014removing} and Hamiltonian quaternions \cite{2011Rain}. Deep-learning-based methods take advantages of deep neural networks to remove undesired snow in images, such as DesnowNet \cite{liu2018desnownet}, CGANs \cite{li2019single}, JSTASR \cite{chen2020jstasr}, HDCWNet \cite{chen2021all} and DDMSNet \cite{zhang2021deep}.
\IEEEpubidadjcol

Despite these great efforts, there are still critical issues to be further addressed. The existing works usually retain a large number of parameter size for better visual qualities, but inevitably, their computational workloads are also remarkably increased. This fact limits their applications in real-world scenarios. Besides, the repairing traces still remain in their results, as shown in the red traffic signs of Fig. \ref{Figure1}(b)(c). Therefore, it is essential to design a lightweight but high-efficient network for this task.

\begin{figure}[ht]
	\includegraphics[width=0.485\textwidth]{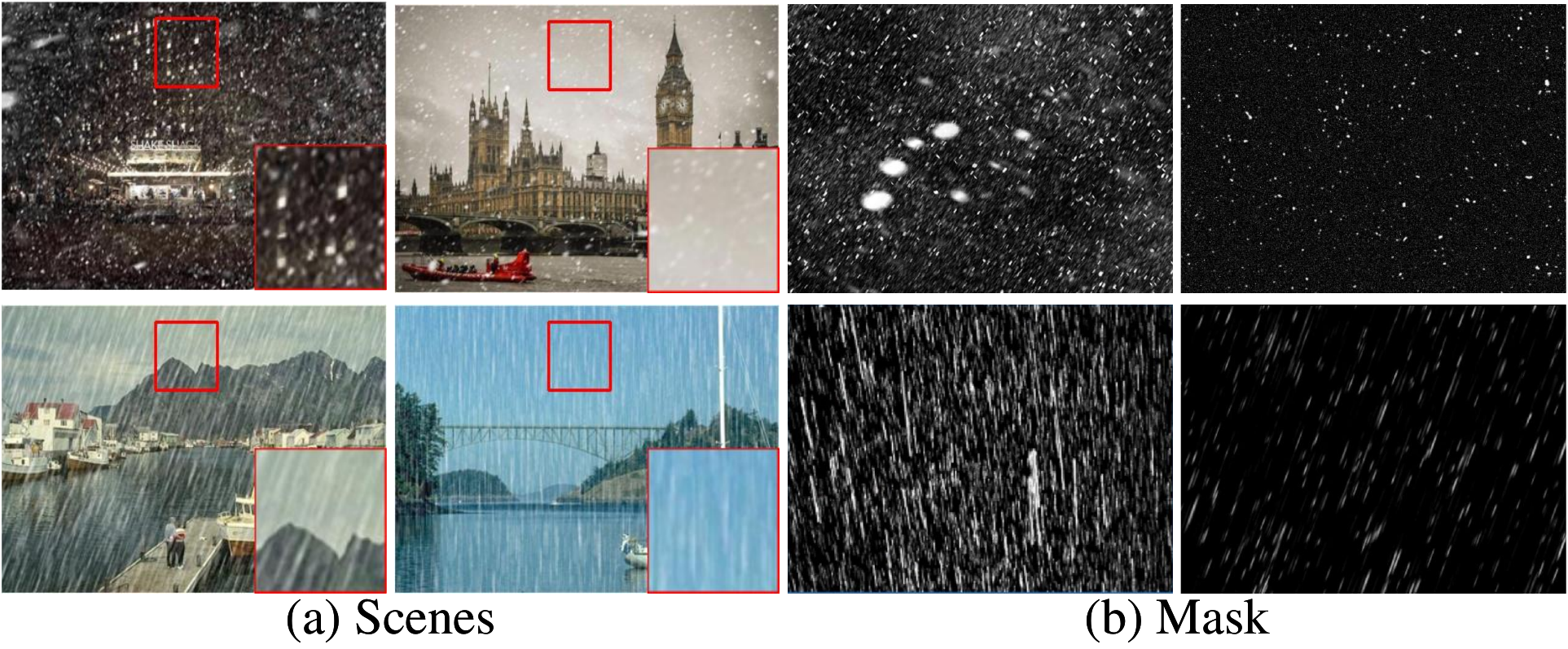}
	\centering
	\caption{Examples of snowy and rainy images. (a) typical snowy and rainy scenes\cite{liu2018desnownet,fu2017removing}. (b) typical snowy and rainy masks \cite{liu2018desnownet,yang2017deep}.}
	\label{Figure2}
\end{figure}

It is noted that existing rain removal methods cannot well address the snow problem due to their apparent visual differences. From Fig. \ref{Figure2}(a), the rain drops and streaks are densely distributed while the snowflakes vary in patterns. The image backgrounds are also more sensible for rainy images. The snowy backgrounds are seriously obscured at different degrees even in the same scene. These differences make it difficult to use existing rain removal methods ({\it e.g.} \cite{yue2021semi,jiang2021rain}) or lightweight methods ({\it e.g.} \cite{cai2019dual, wang2020model}) to process snowy images. How to locate snow areas and recover clean images are still important challenges in snow removal.

To solve the above problems, we propose a lightweight architecture called Laplace-prior-guided Mask Query Transformer (LMQFormer). We observe and demonstrate that Laplace operator can remove redundant information while preserving high-frequency snow edge information. Thus, we take advantage of the lightweight Vector Quantised Variational AutoEncoder (VQVAE) and combine it with the Laplace operator to design a Laplace-VQVAE sub-network. This sub-network generates a coarse mask of snow area, which can be treated as a prior, namely Laplace prior, to guide snow removal. Furthermore, we design a lightweight sub-network called Mask Query Transformer (MQFormer), which uses this coarse mask to obtain high-quality snow-free images with fewer parameters. A Duplicated Mask Query Attention (DMQA) is designed to integrate the coarse mask and concentrate the MQFormer to snow areas.

With this lightweight design, our LMQFormer attains the state-of-the-art visual performance with the smallest parameter size, as demonstrated in Fig. \ref{Figure1}(d)(e). The main contributions are summarized as follows:

\begin{itemize}
\item 
We propose a Laplace-VQVAE sub-network to generate a coarse mask as the prior of snow areas. The effectiveness and low entropy of this Laplace prior inspire us to design a lightweight network without training on masks.

\item 
We design an MQFormer sub-network with DMQA modules for snow removal. The DMQA module employs duplicated mask queries to effectively combine the coarse mask, thus the MQFormer can focus on snow areas and avoid over-enhancement on other regions.

\item 
Extensive experiments on popular benchmarks demonstrate that the proposed LMQFormer network has superior performance and high robustness. Our method also outperforms most state-of-the-arts in terms of processing speed.

\end{itemize}

\section{Related Work}

\subsection{Traditional Snow Removal Methods}
Prior knowledge has been used to guide traditional methods for snow removal. Bossu \textit{et al.} \cite{2011Rain} employed MoG to separate the foreground and used HOG features to detect snow from foreground and recover clean images. Wang \textit{et al.} \cite{2017A} combined image decomposition with dictionary learning to model a three-layer hierarchical snow removal scheme. Zheng \textit{et al.} \cite{zheng2013single} exploited the difference between snow and background and used the multi-guided filter to remove snow. Pei \textit{et al.} \cite{pei2014removing} extracted snow features on saturation and visibility in HSV color space for snow area detection and removal. Rajderkar \textit{et al.} \cite{rajderkar2013removing} applied bilateral filter to decompose a snowy image into low frequency and high frequency parts. Then they decomposed the high frequency part into snowy foreground and clean background by using dictionary learning and sparse coding. Voronin \textit{et al.} \cite{2019Rain} employed anisotropic gradients of Hamiltonian quaternions to detect snow areas and recover clean images. Although these methods have certain effects, they only modeled specific characteristics of snow, which led to poor generalization ability in real-world scenarios.

\subsection{Deep-learning-based Snow Removal Methods}
Recently, deep learning has been successfully applied in single image snow removal task. Liu \textit{et al.} \cite{liu2018desnownet} proposed the first deep-learning-based method called DesnowNet. They applied a two-stage network to learn the mapping from snowy images to mask and recover clean images. Li \textit{et al.} \cite{li2019single} used Generative Adversarial Network (GAN) to recover clean images. Chen \textit{et al.} \cite{chen2020jstasr} proposed a JSTASR model to generate three different snow masks with differentiable dark channel prior layer and guide image recovery with these masks. Li \textit{et al.} \cite{li2020all} created architectural search and proposed all-in-one network for snow, rain and fog removal. Chen \textit{et al.} \cite{chen2021all} proposed HDCWNet, which removed snow with hierarchical dual-tree complex wavelet representation and contradict channel loss. Zhang \textit{et al.} \cite{zhang2021deep} designed DDMSNet and introduced semantic and geometric images as prior knowledge to guide snow removal. Jaw  \textit{et al.} \cite{chen2022learning} presented an efficient and highly modularized network called DesnowGAN. Quan \textit{et al.} \cite{quan2023image} proposed an Invertible Neural Network (INN) to predict a latent image and a snowflake layer for snow removal. Chen \textit{et al.} \cite{jaw2021desnowgan} used a two-stage knowledge distillation learning to recover all bad-weather images, which is called TKL in this paper. In summary, these methods achieve promising performances in snow removal but at costs of a large parameter size. Besides, their recovered images may remain repairing traces, as shown in Fig. \ref{Figure1}(b)(c). 

Different from these methods, we aim to explore a prior of coarse mask to guide a lightweight but high-efficient network. The proposed model also generates images with high visual qualities which will be validated in experimental results.

\subsection{Other Image Denoising Tasks}
There are some other similar denoising tasks, including rain removal ({\it e.g.} \cite{fu2021successive}), haze removal ({\it e.g.} \cite{nie2022stereo}), low-light enhancement ({\it e.g.} \cite{li2021low}) and underwater image enhancement ({\it e.g.} \cite{jiang2022}). For example, Zhang \textit{et al.} \cite{zhang2019image} proposed a novel single image deraining method called image de-raining conditional generative adversarial network (ID-CGAN). Zhang \textit{et al.} \cite{zhang2022enhanced} proposed an Enhanced Spatio-Temporal Integration Network (ESTINet) to exploit spatio-temporal information for rain streak removal. Agrawal \textit{et al.} \cite{9385131} presented a single image dehazing method based on a superpixel, nonlinear transformation. Wang \textit{et al.} \cite{wang2022low} introduced a normalizing flow model for low-light enhancement. Jiang \textit{et al.} \cite{jiang2021underwater} exploited the potential of lightweight network that benefits both effectiveness and efficiency for underwater image enhancement.

\begin{figure}[ht]
	\includegraphics[width=0.485\textwidth]{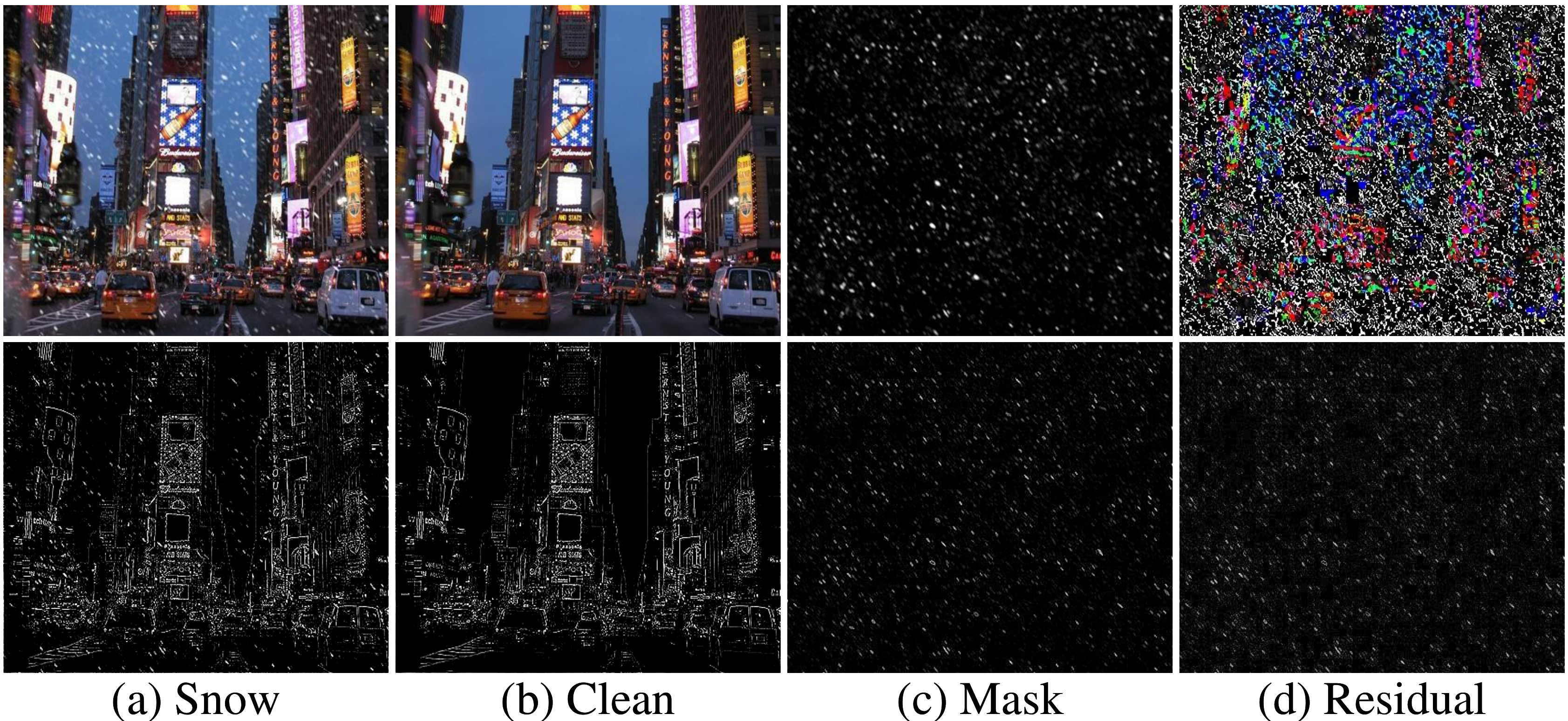}
	\centering
	\caption{Snow images processed by Laplace operator. (a) a snowy image before and after being processed by Laplace operator; (b) the clean background images of (a); (c) the snow masks of (a); (d) the residuals of (a)-(b).}
	\label{Figure3}
\end{figure}

\section{Methodology}

\subsection{Problem Statement}
As discussed in Section II, deep-learning-based snow removal methods take advantage of CNNs to recover clean images. Their deep neural networks learns a mapping from the input snowy image $I_{\rm snow}$ and its output $I_{\rm clean}$:
\begin{equation}
    I_{\rm clean}={\rm F_m}(I_{\rm snow}; \theta),
\end{equation}
where $\rm F_m(\cdot)$ represents a deep neural network for mapping, $\theta$ represents the parameters of $\rm F_m(\cdot)$. Considering the complicated patterns of snowflakes, existing methods tend to employ complicated networks with a large parameter size. Although they achieve successes in recovering clean images under snow, their parameter volumes limit their use in practice scenarios {\it e.g.} surveillance, videos, etc. The snow removal task is still calling for an effective model with a low parameter size and a high processing speed.

To address this issue, we revisit the physical model of snowy image as shown in \cite{liu2018desnownet}:
\begin{equation}
I_{\rm snow}=R \odot M+C \odot(1-M),
\end{equation}where $R$, $C$ and $M$ are the chromatic aberration map, the latent clean image and the mask image of $I_{\rm snow}$, respectively. This decomposition inspires us to design a mask-based snow removal approach. However, the snow mask is not available in practical use. Instead, we interpret a coarse mask and further utilize it to form a unified prior that benefits both snow location and removal. This task is thus defined as:
\begin{equation}
    \begin{aligned}
      &I_{\rm prior}={\rm G_n}(I_{\rm snow}; \theta_{\rm A}),
    \\ &I_{\rm clean}=I_{\rm snow} - {\rm F_n}(I_{\rm snow}, I_{\rm prior}; \theta_{\rm B}),      
    \end{aligned}
    \label{3}	
\end{equation}
where $I_{\rm prior}$ represents this unified prior. $\rm G_n(\cdot)$ and $\rm F_n(\cdot)$ represent the sub-networks to generate $I_{\rm prior}$ and recover clean images with $I_{\rm snow}$ and $I_{\rm prior}$, respectively.

Based on the above analyses, we attempt to construct a unified prior of snow via an interpreted coarse mask and further utilize it in recovering clean images. As shown in Eq. (\ref{3}), the functions of our prior are twofold. First, it is utilized to coarsely locate snow areas. Second, it is combined with $I_{\rm snow}$ to estimate the residuals between the input and output of our model. With a guidance of this prior, we can achieve better snow removal results under the lightweight requirement.

\subsection{Laplace Prior}
This paper calculates the $I_{\rm prior}$ with the sub-network $\rm G_n(\cdot)$ and Laplace operator. It is commonly known that lower entropy of data makes ease of feature extraction of neural network, that is, the network is easier to learn latent rules of data. Here we define an optimization of the relative entropy between $I_{\rm prior}$ and $I_{\rm snow}$:
\begin{equation}
    \begin{aligned}
    & \min_{q_\theta} D_{KL} \left(q_\theta(I_{\rm prior} \mid I_{\rm snow})\|p(I_{\rm prior})\right)
    \\  s.t. \quad I_{\rm prior} &\sim q_\theta(I_{\rm prior} \mid I_{\rm snow}) \Longleftrightarrow I_{\rm prior}={\rm G_n}(I_{\rm snow}; \theta_{\rm A}),
    \end{aligned}
\end{equation}
where $p(I_{\rm prior})$ and $q_\theta(I_{\rm prior} \mid I_{\rm snow})$ represent the prior and posterior probabilities of $I_{\rm prior}$, respectively. This minimization process can help to design the network $\rm G_n(\cdot)$ and its learning parameters $\theta_{\rm A}$. 

In the designing of network $\rm G_n(\cdot)$, we amplify the snow features and eliminate the irrelevant information to reduce $D_{KL}$. This can be achieved with Laplace operator:
\begin{equation}
    I_{\rm lapsnow} = \nabla^{2}I_{\rm snow}.
\end{equation}
Correspondingly, the clean image $I_{\rm clean}$ outputs $I_{\rm lapclean}$ with Laplace operator. From Fig. \ref{Figure3}, it is extremely easy to distinguish $I_{\rm snow} - I_{\rm clean}$ ({\it i.e.} the fourth image in the first row) and the snow mask ({\it i.e.} the third image in the first row). However, after being processed by Laplace operator, the corresponding images are quite similar, as shown in the second row of Fig. \ref{Figure3}(c)(d). Thus, we are allowed to use $I_{\rm lapsnow} - I_{\rm lapclean}$ as a coarse mask for snow removal.

With Laplace operator, $\rm G_n(I_{\rm snow}; \theta_A)$ is replaced by $\rm G_L(I_{\rm snow}; \theta_A)$. Its training process can be expressed as:
\begin{equation}
    \min _{{\rm G_L}(\cdot)}\left\|{\rm G_L}(I_{\rm snow}; \theta_{\rm A}) - (I_{\rm lapsnow} - I_{\rm lapclean})\right\|_2 ,
    \label{6}
\end{equation}
where we attempt to train a Laplace prior to approximate the coarse mask without explicit knowledge of Fig. \ref{Figure3}(c).

\subsection{Lightweight Design}
With Laplace operator, we can approximate the coarse mask as a prior, as shown in Eqs. (\ref{3}) and (\ref{6}) and further utilize it to guide the recovery process, as shown in Eq. (\ref{3}). This operation allows us to compute more efficiently with low entropy data. The complete model is designed as a lightweight network, which is benefited from the following methods. First, we introduce spatial attention and Codebook \cite{van2017neural} for $\rm G_L(\cdot)$ to effectively extract spatial features for the coarse mask. Second, we introduce Mask Query Transformer Module (MQTM) to make $\rm F_n(\cdot)$ concentrate on snow areas, thus, we call our recovery network as $\rm F_{\rm MQ}(\cdot)$. Third, we design an efficient framework consisting of parallel transformer and convolutional encoders as well as a hybrid decoder to obtain more representative features of snow at low scales. This framework greatly reduce computational workloads while remaining high-performance. By considering all these issues, our complete model is designed as:
\begin{equation}
    \begin{aligned}
    &I_{\rm clean}= I_{\rm snow} - {\rm F_{MQ}}(I_{\rm snow}, {\rm G_{L}}(I_{\rm snow}; \theta_{\rm A}); \theta_{\rm B}),
    \end{aligned}
\end{equation}
where $\rm G_L(\cdot)$ and $\rm F_{\rm MQ}(\cdot)$ represent the coarse mask generator and the recovery network, respectively.

\begin{figure*}[ht]
    \includegraphics[width=1\textwidth]{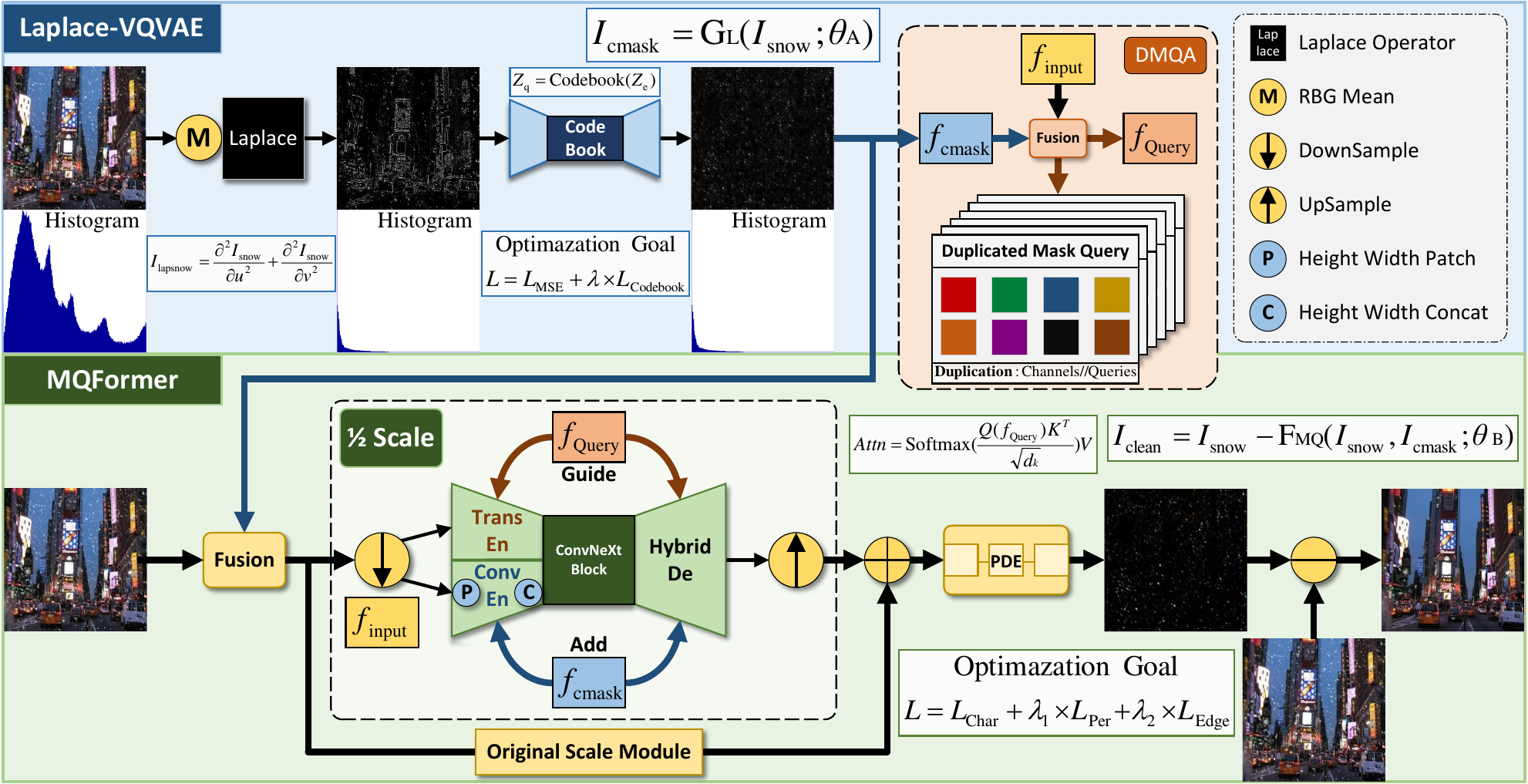}
    \centering
    \caption{The proposed LMQFormer with two sub-networks: Laplace-VQVAE and MQFormer. The Laplace-VQVAE extracts a prior of the coarse mask while the MQFormer recover clean images with the Laplace prior. $f_{\rm input}$ and $f_{\rm cmask}$ are fused to get $f_{\rm Query}$ for the calculation of $Attn$ in DMQA. Due to Laplace operation and the attention-based design, our network is lightweight but high-efficient.}
    \label{Figure4}
\end{figure*}

\begin{figure}[ht]
    \includegraphics[width=0.485\textwidth]{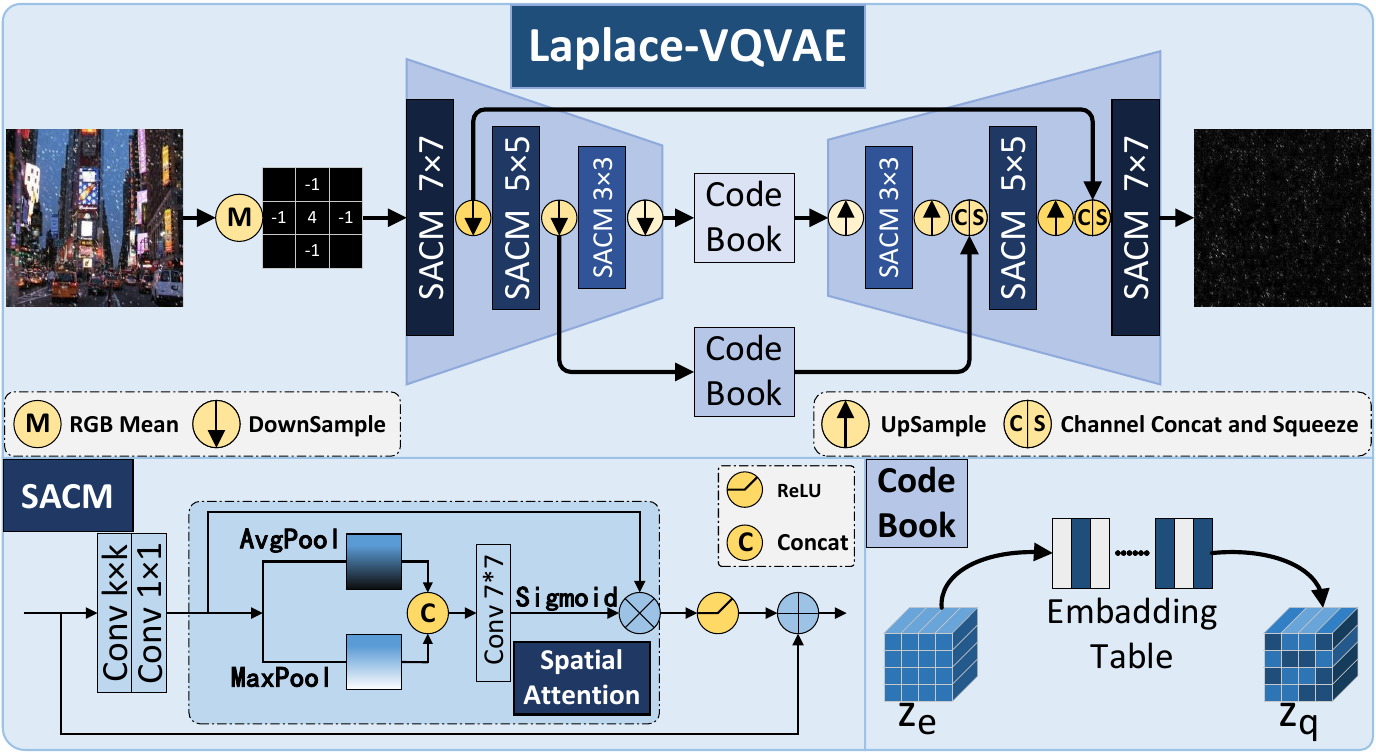}
    \centering
    \caption{The proposed Laplace-VQVAE with SACM and Codebook. The input image is filtered by a Laplace operator, and then processed by a multi-scale encoder-decoder. At each layer of different scales, SACM learns spatial characteristics of snow by different kernel sets and spatial attention. The two Codebooks represent latent code z of the coarse mask.}
    \label{Figure5}
\end{figure}

\section{Network Design}
As shown in Fig. \ref{Figure4}, we realize $\rm G_L(\cdot)$ and $\rm F_{\rm MQ}(\cdot)$ by two sub-networks: Laplace-VQVAE and MQFormer. They formulate a complete Laplace-prior-guided Mask Query Transformer (LMQFormer). In Laplace-VQVAE, Laplace operator reduces the entropy of snow images and makes it easy to build a lightweight sub-network to generate the coarse mask. Then a DMQA module is utilized to guide MQFormer to pay more and concentrate attentions to snow areas. In MQFormer, complex module calculations are carried out at low scales. An efficient framework consisting of parallel transformer and convolutional encoders as well as a hybrid decoder are designed to learn diversity representation of features. These designs guarantee the effectiveness of our model with reduced parameters.

\subsection{Laplace-VQVAE}
In our network, we use the VQVAE-based \cite{van2017neural} design instead of VAE-based network which is mainly due to the lightweight property of VQVAE. Based on this, we design a lightweight sub-network called Laplace-VQVAE to generate the coarse mask. As shown in Fig. \ref{Figure5}, we obtain gray images and then filtered by Laplace operator. Then, we design three downsample layers to obtain multi-scale features and introduce Codebook at two low scales to represent the coarse mask for latent code $z$. We directly hop features at one high scale to compensate for the loss of mask information during downsampling. Spatial Attention Convolution Module (SACM) is designed to preserve important spatial information of snow. We inherit the lightweight inspiration from \cite{zamir2021multi} to reduce parameters, and utilize spatial attention from \cite{woo2018cbam} and different convolution kernels to improve performance. Finally, we get the coarse mask $I_{\rm prior}$ as $I_{\rm cmask}$. Based on this design, our Laplace-VQVAE has only 157K parameters with 24 channels.

\begin{figure*}[ht]
    \includegraphics[width=0.9\textwidth]{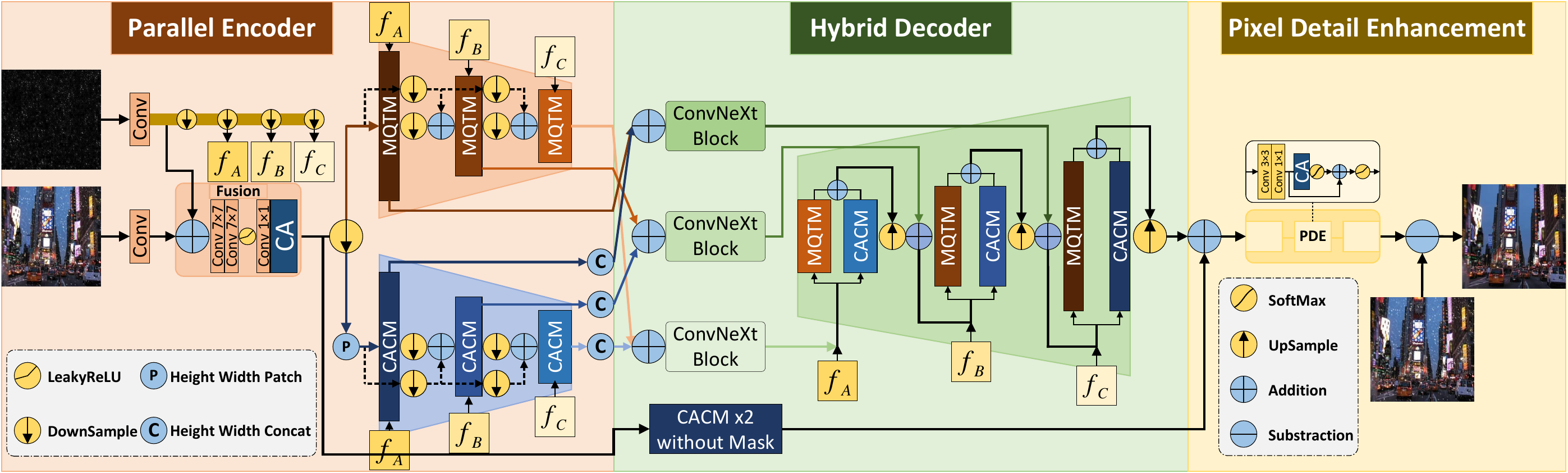}
    \centering
    \caption{The proposed MQFormer consisting of a transformer encoder, a convolutional encoder and a hybrid decoder. \(f_A\), \(f_B\) and \(f_C\) represent stepwise downsampling of the coarse mask features. The coarse mask and input snow image are first fused by the Fusion block, then encoded by two parallel encoders. Each scale of encoded features is fused by a ConvNeXt block, then decoded by a hybrid decoder to get learned snow features.}
    \label{Figure6}
\end{figure*}

\begin{figure}[ht]
    \includegraphics[width=0.45\textwidth]{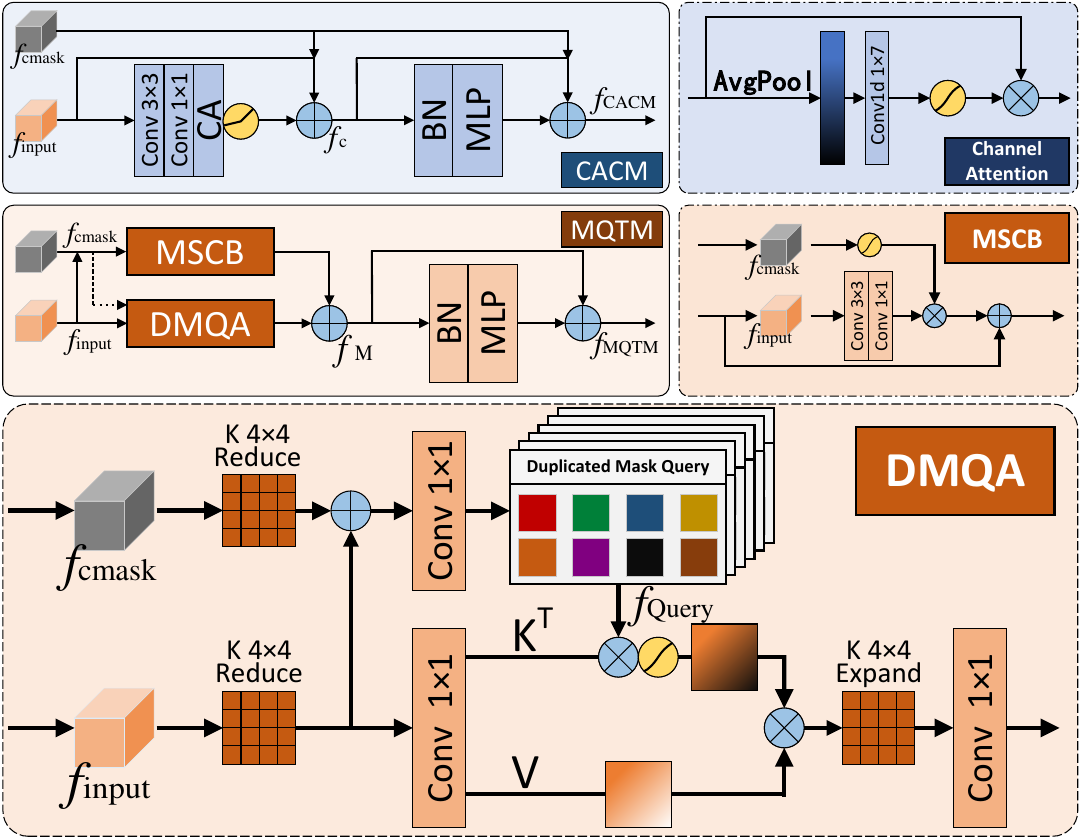}
    \centering
    \caption{The details of MQTM and CACM in our parallel encoders and hybrid decoder. In MQTM, both $f_{\rm cmask}$ and $f_{\rm input}$ are sent into MSCB and DMQA. In CACM, $f_{\rm cmask}$ is directly added to the features to prevent information loss.}
    \label{Figure7}
\end{figure}

VAE \cite{2014Auto} learns the latent variable $z$ through Evidence Lower Bound (ELBO):
\begin{equation}
	E\left[\log p(I_{\rm snow} \!\mid\! z)\right] \!-\! D_{KL}\left(q(z \!\mid\! I_{\rm snow}) \| p(z)\right),
\end{equation}
where $p(z)$ and $q(z \!\mid\! I_{\rm snow})$ are prior and posterior probabilities of $z$, respectively. VQVAE introduces the Codebook of embedding table to generate the vector $z_q$. It replaces the output of encoder $z_e$ with its nearest embedding in the table. Through this operation, the original probability distribution of $q(z \!\mid\! I_{\rm snow})$ becomes into a $K_d$-dimensional one-hot form, which corresponds to the probability of the embedding table in Codebook. Then $D_{KL}$ term becomes a constant:
\begin{equation}
	\begin{aligned}
		& D_{KL} \left(q(z \!\mid\! I_{\rm snow}) \| p(z)\right) \hfill
		\\ &= 1 \!\cdot\! \log \left(\frac{1}{\frac{1}{K_d}}\right)+({K_d}\!-\!1) \!\cdot\! 0 \!\cdot\! \log \left(\frac{0}{\frac{1}{K_d}}\right) \!=\!\log {K_d} \hfill.
	\vspace{-1em}
	\end{aligned}
\end{equation}The Codebook optimization goal can be expressed as:
\begin{equation}
	\mathcal{L}^{k}_{\rm Codebook}=\sum_{k}^{2}\sum_{j}^{n_{i}}\left\|z_{i, j}-e_{i}\right\| ,
\end{equation} 
where \(k\) reprensents different layers, \(e_{i}\) represents vectors in Codebook, $i,j$ represent the index of latent code $z$. Inspired by VQVAE, we exploit Codebook at low scales to model important features of mask.

\subsection{MQFormer}
Fig. \ref{Figure6} shows the details of our proposed MQFormer. We design an efficient framework consisting of two parallel encoders, a hybrid decoder and modified ConvNeXtBlock \cite{liu2022convnet} to learn multi-scale features at low scales. Moreover, Pixel Detail Enhancement (PDE) learn further details on the original scale. The coarse mask obtains $f_A$, $f_B$ and $f_C$ through continuous downsampling, then uses them to guide encoders and decoder to learn related snow features.

The two parallel encoders and the hybrid decoder utilize Channel Attention Convolution Modules (CACMs) and Mask Query Transformer Modules (MQTMs) to realize our lightweight design. As shown in Fig. \ref{Figure7}, the CACM module adds the coarse mask to avoid information loss. It utilized a Channel Attention (CA) with 7 parameters only due to 1D convolution. This process can be formulated as follows:
\begin{equation}
	\begin{aligned}
		& f_{\rm CA} = {\rm LReLU}[{\rm CA}({\rm Conv}(f_{\rm input}))], \\
		& f_{\rm C} = f_{\rm CA} + f_{\rm input} + f_{\rm cmask}, \\
		& f_{\rm CACM} = f_{\rm C} + {\rm BN}[{\rm MLP}(f_{\rm C})] + f_{\rm cmask},
	\end{aligned}
\end{equation}
where ${\rm LReLU[\cdot]}$ represents LeakyReLU activation function, ${\rm BN[\cdot]}$ represents BatchNorm.

The MQTM module uses two parallel blocks for feature extraction: DMQA and Mask Spatial Convolution Block (MSCB). In DMQA, the coarse mask $f_{\rm cmask}$ is added to Query and reduced to a specific query number. Then, the attention is calculated at ¼ scales of $f_{\rm input}$ through 4$\times$4 kernels. After that, duplicated masked query is performed at low scale and further expanded to the original scale with 4$\times$4 convolutions. In MSCB, the coarse mask is multiplied in after a Softmax operation to compensate for detail loss in the downsampling process of DMQA. Besides, MQTM has the same architecture of BatchNorm and MLP as CACM. Finally, we obtain a fused feature $f$ with a combination of DMQA and MSCB. This process can be formulated as follows:
\begin{equation}
	\begin{aligned}
		& K, V \leftarrow {\rm CS}[{\rm Conv}_{1\times1}({\rm RD}(f_{\rm input}))], \\
		& f_{\rm Query} \leftarrow {\rm CD}[{\rm Conv}_{1\times1}({\rm RD}(f_{\rm input}) + {\rm RD}(f_{\rm cmask}))], \\
		& f_{\rm DMQA} = {\rm EP}({\rm Softmax}\left(\frac{f_{\rm Query} K^T}{\sqrt{d_k}}\right) V), \\
		& f_{\rm MSCB} = f_{\rm input} + {\rm Conv}(f_{\rm input}) \odot {\rm Softmax}(f_{\rm cmask}), \\
		& f_{\rm M} = f_{\rm MSCB} + f_{\rm DMQA}, \\
		& f_{\rm MQTM} = f_{\rm M} + {\rm BN}[{\rm MLP}(f_{\rm M})],
	\end{aligned}
\end{equation}where \({\rm CS[\cdot]}\) and \({\rm CD[\cdot]}\) represent channel-wise split and duplication, respectively, ${\rm RD(\cdot)}$ and ${\rm EP(\cdot)}$ represent reduction and expansion by $4\times4$ kernel, respectively. Since LayerNorm barely contributes to visual quality with a low processing speed, we replace it with BatchNorm. We set the ratio of MLP to 4 in MQFormer. Finally, we subtract the learned residual snow map from $I_{\rm snow}$ to get $I_{\rm clean}$.

\begin{figure*}[ht]
    \includegraphics[width=0.9\textwidth]{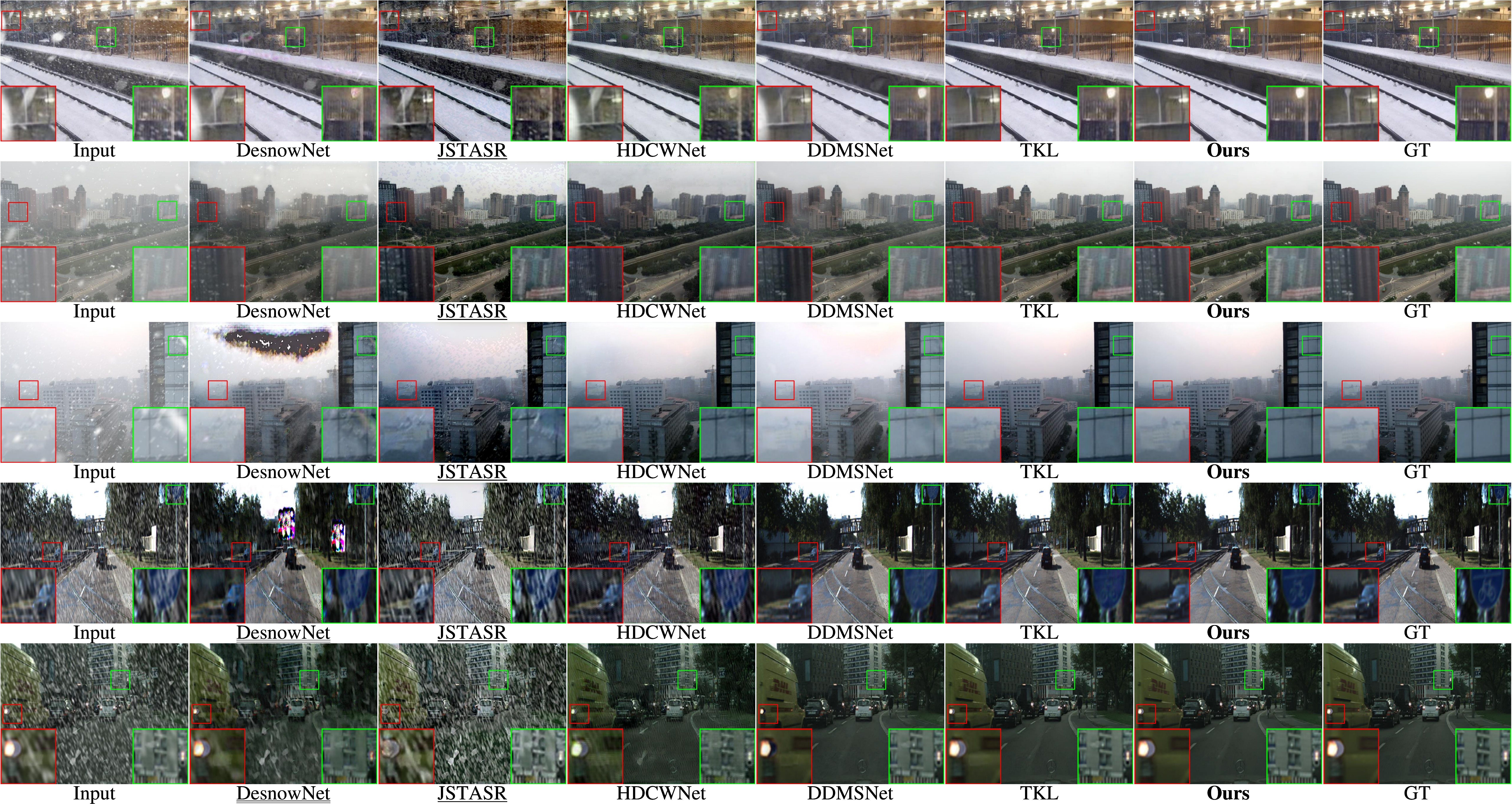}
    \centering
    \caption{Qualitative results on synthetic datasets. From top to bottom rows: Snow100K, SRRS, CSD, SnowKitti2012 and SnowCityScapes. Please zoom in for better visual quality.}
    \label{Figure8}
\end{figure*}

\subsection{Training and Loss Functions}
Our proposed LMQFormer is trained with two steps. Firstly, Laplace-VQVAE is optimized by Adam and trained for 100 epochs. Secondly, the weight of Laplace-VQVAE is fixed to train MQFormer. MQFormer is optimized by AdamW and trained for 400 epochs. In addition, all sub-networks are trained on randomly cropped 256\(\times\)256 patches with data augmentation of randomly horizontal and vertical flips. The validation dataset is cropped to 256\(\times\)256 in the center. Two learning rates are steadily decreased from the initial learning rate of \(2\times10^{-4}\) to \(1\times10^{-6}\) using the cosine annealing algorithm. 

Laplace-VQVAE is trained with the squared penalty of Mean Square Error (MSE) \(\mathcal{L}_{\rm MSE}\) and Codebook Loss \(\mathcal{L}_{\rm Codebook}\). The total optimization can be expressed as:
\begin{equation}
	\mathcal{L}_{\rm LaplaceVQVAE} = \mathcal{L}_{\rm MSE} + \lambda\times\mathcal{L}_{\rm Codebook},
\end{equation}where \(\lambda\)=0.25 is a hyper-parameter of Codebook. 

MQFormer is trained with three losses: Charbonnier Loss \(\mathcal{L}_{\rm char}\) \cite{lai2018fast}, Perceptual Loss \(\mathcal{L}_{\rm per}\) of VGG19 \cite{johnson2016perceptual} and Edge Loss \(\mathcal{L}_{\rm edge}\). Edge Loss can be expressed as:
\begin{equation}
	\mathcal{L}_{\rm edge}=\frac{1}{N}\sum_{i=1}^{N}
	\sqrt{\left\|\Delta\left({I_{\rm pred}}\right)-\Delta({I_{\rm clean}})\right\|^{2}+\varepsilon^{2}},
\end{equation} 
where \(\varepsilon\)=\(10^{-5}\) represents hyper-parameter constant. The total optimization can be expressed as:
\begin{equation}
	\mathcal{L}_{\rm MQFormer} = \mathcal{L}_{\rm char} + \lambda_{1}\times\mathcal{L}_{\rm per} + \lambda_{2}\times\mathcal{L}_{\rm edge},
\end{equation} 
where typically, \(\lambda_{1}\)=0.1 and \(\lambda_{2}\)=0.05.

\begin{table*}[ht]
\caption {Quantitative results on synthetic datasets. {\color[HTML]{FF0000} Red} and {\color[HTML]{0000FF} blue} indicate the best and the second-best results, respectively.}
\label{Table1}
\centering
\renewcommand
\arraystretch{1.8}
\scalebox{1}{
    \begin{tabular}{cc|cccccc}
    \toprule
    \multicolumn{1}{c|}{\textbf{\begin{tabular}[c]{@{}c@{}}Dataset\\ (trainset, testset)\end{tabular}}}                           & \textbf{IQA}    & \textbf{\begin{tabular}[c]{@{}c@{}}DesnowNet\\ (TIP'18)\end{tabular}} & \textbf{\begin{tabular}[c]{@{}c@{}}JSTASR\\ (ECCV'20)\end{tabular}} & \textbf{\begin{tabular}[c]{@{}c@{}}HDCWNet\\ (ICCV'21)\end{tabular}} & \textbf{\begin{tabular}[c]{@{}c@{}}DDMSNet\\ (TIP'21)\end{tabular}} & \textbf{\begin{tabular}[c]{@{}c@{}}TKL\\ (CVPR'22)\end{tabular}} & \textbf{Ours}                        \\ \midrule

    \multicolumn{1}{c|}{}                                                                                                     & PSNR(↑)/SSIM(↑) & 23.125/0.788                                                          & \uline{18.648}/\uline{0.554}                                                  & 21.296/0.640                                                         & 29.058/0.891                                                        & {\color[HTML]{0000FF} 30.507}/{\color[HTML]{0000FF} 0.895}                              & {\color[HTML]{FF0000} 31.883}/{\color[HTML]{FF0000} 0.917}  \\
    \multicolumn{1}{c|}{\multirow{-2}{*}{\textbf{\begin{tabular}[c]{@{}c@{}}Snow100K\\ (10000, 1200)\end{tabular}}}}      & MAE(↓)/LPIPS(↓) & {\color[HTML]{0000FF} 100.387}/0.257                                  & \uline{129.854}/\uline{0.437}                                                 & 156.553/0.405                                                        & 168.364/0.103                                                       & 106.921/{\color[HTML]{0000FF} 0.092}                                                     & {\color[HTML]{FF0000} 93.999}/{\color[HTML]{FF0000} 0.085}  \\ \hline

    \multicolumn{1}{c|}{}                                                                                                     & PSNR(↑)/SSIM(↑) & 21.307/0.835                                                          & \uline{21.593}/\uline{0.770}                                                  & 24.165/0.823                                                         & 25.967/0.932                                                        & {\color[HTML]{0000FF} 29.374}/{\color[HTML]{0000FF} 0.934}                              & {\color[HTML]{FF0000} 31.040}/{\color[HTML]{FF0000}0.964}  \\
    \multicolumn{1}{c|}{\multirow{-2}{*}{\textbf{\begin{tabular}[c]{@{}c@{}}SRRS\\ (8000, 1200)\end{tabular}}}}           & MAE(↓)/LPIPS(↓) & {\color[HTML]{0000FF} 119.795}/0.290                                  & \uline{178.618}/\uline{0.245}                                                 & 155.091/0.263                                                        & 136.799/0.058                                                       & 128.530/{\color[HTML]{0000FF} 0.050}                                                    & {\color[HTML]{FF0000} 119.230}/{\color[HTML]{FF0000} 0.025} \\ \hline

    \multicolumn{1}{c|}{}                                                                                                     & PSNR(↑)/SSIM(↑) & 20.632/0.777                                                          & \uline{20.628}/\uline{0.705}                                                  & 28.669/0.892                                                         & 24.976/0.905                                                        & {\color[HTML]{FF0000} 33.351}/{\color[HTML]{0000FF} 0.954}                              & {\color[HTML]{0000FF} 32.643}/{\color[HTML]{FF0000} 0.963}  \\
    \multicolumn{1}{c|}{\multirow{-2}{*}{\textbf{\begin{tabular}[c]{@{}c@{}}CSD\\ (8000, 1200)\end{tabular}}}}            & MAE(↓)/LPIPS(↓) & 152.851/0.309                                                         & \uline{193.095}/\uline{0.410}                                                 & 131.587/0.118                                                        & 130.517/0.084                                                       & {\color[HTML]{FF0000} 121.548}/{\color[HTML]{0000FF} 0.035}                             & {\color[HTML]{0000FF} 121.900}/{\color[HTML]{FF0000} 0.029} \\ \hline
    \multicolumn{1}{c|}{}                                                                                                     & PSNR(↑)/SSIM(↑) & \uuline{16.487}/\uuline{0.682}                                                    & \uline{18.470}/\uline{0.532}                                                  & 22.348/0.644                                                         & 30.520/0.934                                                        & {\color[HTML]{0000FF} 31.354}/{\color[HTML]{0000FF} 0.939}                              & {\color[HTML]{FF0000} 33.145}/{\color[HTML]{FF0000} 0.959}  \\

    \multicolumn{1}{c|}{\multirow{-2}{*}{\textbf{\begin{tabular}[c]{@{}c@{}}SnowKitti2012\\ (4500, 1200)\end{tabular}}}}  & MAE(↓)/LPIPS(↓) & {\color[HTML]{0000FF} \uuline{96.785}}/\uuline{0.353}                            & \uline{143.152}/\uline{0.407}                                                 & 161.180/0.344                                                        & 112.250/{\color[HTML]{0000FF} 0.053}                                                       & 97.869/0.058                                                     & {\color[HTML]{FF0000} 81.847}/{\color[HTML]{FF0000} 0.051}  \\ \hline
    \multicolumn{1}{c|}{}                                                                                                     & PSNR(↑)/SSIM(↑) & \uuline{21.020}/\uuline{0.705}                                                    & \uline{19.047}/\uline{0.550}                                                  & 25.236/0.792                                                         & 33.371/0.956                                                        & {\color[HTML]{0000FF} 35.691}/{\color[HTML]{0000FF} 0.967}                              & {\color[HTML]{FF0000} 39.172}/{\color[HTML]{FF0000} 0.984}  \\
    
    \multicolumn{1}{c|}{\multirow{-2}{*}{\textbf{\begin{tabular}[c]{@{}c@{}}SnowCityScapes\\ (6000, 1200)\end{tabular}}}} & MAE(↓)/LPIPS(↓) & {\color[HTML]{0000FF} \uuline{79.821}}/\uuline{0.343}                             & \uline{129.134}/\uline{0.419}                                                 & 188.574/0.255                                                        & 105.535/0.031                                                       & 98.991/{\color[HTML]{0000FF} 0.018}                                                  & {\color[HTML]{FF0000} 67.639}/{\color[HTML]{FF0000}0.010}  \\ \hline

    \multicolumn{1}{c|}{}  & PSNR(↑)/SSIM(↑) & 21.929/0.737                                                          & \uline{17.925}/\uline{0.520}                                                        & 20.883/0.618                                                         & 27.018/0.849                                                        & {\color[HTML]{0000FF} 28.368}/{\color[HTML]{0000FF} 0.863}                              & {\color[HTML]{FF0000} 29.705}/{\color[HTML]{FF0000} 0.890}\\
    \multicolumn{1}{c|}{\multirow{-2}{*}{\textbf{\begin{tabular}[c]{@{}c@{}}Snow100K-L\\ (10000, 1000)\end{tabular}}}}           & MAE(↓)/LPIPS(↓) & {\color[HTML]{FF0000} 92.829}/0.306                                   & \uline{107.352}/\uline{0.471}                                                       & 143.796/0.429                                                        & 135.532/0.141                                                       & 113.583/{\color[HTML]{0000FF} 0.117}                                                    & {\color[HTML]{0000FF} 103.560}/{\color[HTML]{FF0000} 0.097}  \\

    \midrule
    \multicolumn{2}{c|}{\textbf{Parameters(M)}}                                                                                           & 26.151                                                                & 83.347                                                              & {\color[HTML]{0000FF} 18.958}                                        & 229.454                                                             & 28.712                                                           & {\color[HTML]{FF0000} 2.181}         \\ \hline
    \multicolumn{2}{c|}{\textbf{Runtimes(s)}}                                                                                                   & 2.992                                                                 & 0.356                                                               & 0.141                                                                & 0.554                                                               & {\color[HTML]{0000FF} 0.051}                                     & {\color[HTML]{FF0000} 0.042}         \\ \bottomrule
    \end{tabular}}
\end{table*}

\begin{figure*}[ht]
    \includegraphics[width=1.0\textwidth]{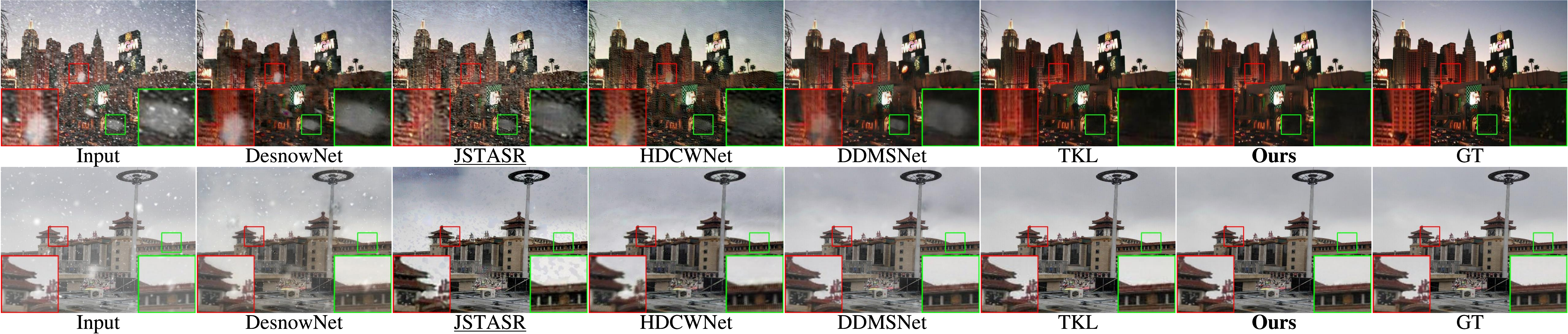}
    \centering
    \caption{Qualitative results on large snowflakes and veiling effect. Please zoom in for better visual quality.}
    \label{Figure9}
\end{figure*}

\begin{figure*}[ht]
    \includegraphics[width=1.0\textwidth]{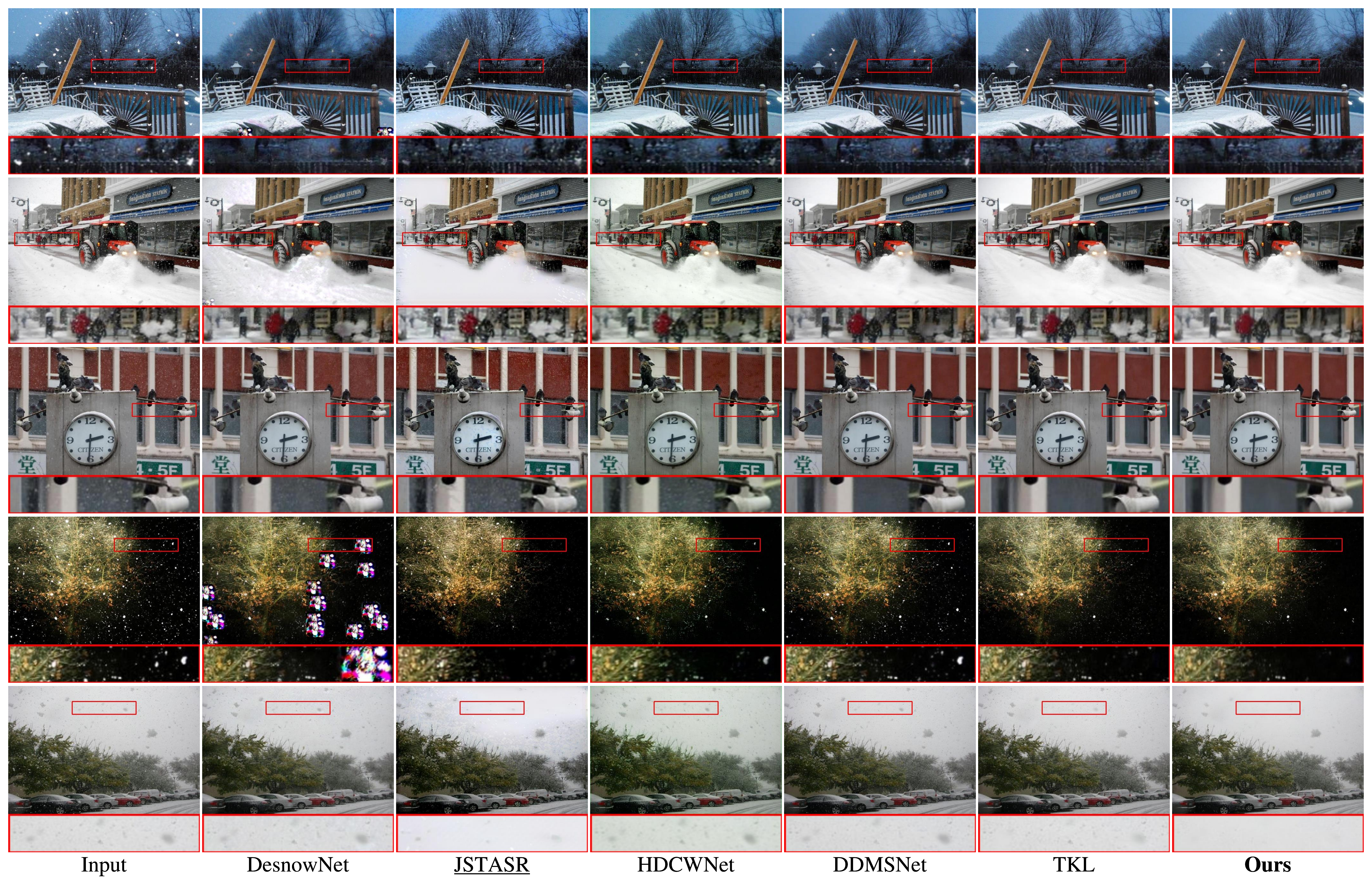}
    \centering
    \caption{Qualitative results on real-world images. Please zoom in for better visual quality.}
    \label{Figure10}
\end{figure*}

\section{Experiments and Analysis}

\subsection{Datasets and Evaluation}

\subsubsection{Synthetic Datasets}
To better evaluate our network, we run it on all available synthetic snow removal datasets. They contain Snow100K \cite{liu2018desnownet}, SRRS \cite{chen2020jstasr}, CSD \cite{chen2021all}, SnowKITTI2012 \cite{zhang2021deep} and SnowCityScapes \cite{zhang2021deep}. Snow100K is the first snow removal dataset, which contains 50000 images in both training and testing sets. SRRS exploits the veiling effect in dataset and contains 12000 images. CSD introduces snow streaks after SRRS with 8000 images in training set and 2000 images in testing set. Besides, SnowKITTI2012 and SnowCityScapes are street view snow removal datasets. SnowKITTI2012 contains 4500 images in training set and 3000 images in testing set. SnowCityScapes contains 6000 images in training set and testing set. Since snow scenarios are very different for each dataset, multiple datasets allow for a more comprehensive assessment of snow removal methods.

\subsubsection{Real-world Dataset}
We employ the Snow100K real-world dataset \cite{liu2018desnownet} to verify the generalization ability of algorithms in the real-world scenarios. This dataset consists of 1329 images with rich real-world scenarios.

\subsubsection{Evaluation}
Each dataset consists of a different number of images. We randomly select a specific number of images from each synthetic dataset for training. As for evaluations, we randomly select 1200 images from each synthetic dataset. We use four popular quality metric for evaluation, including the Peak Signal to Noise Ratio (PSNR) \cite{2006A}, Structural Similarity Index (SSIM) \cite{2004Image}, Mean Absolute Error (MAE) and Learned Perceptual Image Patch Similarity (LPIPS) \cite{zhang2018unreasonable}. All experiments are performed on a server equipped with four NVIDIA RTX TITAN X GPU with 12GB GPU memory, Linux and PyTorch.

\subsubsection{Compared Methods}
We compare our method with state-of-the-art deep-learning-based snow removal methods using publically available codes released by the authors, including DesnowNet \cite{liu2018desnownet} (TIP'18), JSTASR \cite{chen2020jstasr} (ECCV'20), HDWCNet \cite{chen2021all} (ICCV'21), DDMSNet \cite{zhang2021deep} (TIP'21) and TKL \cite{chen2022learning} (CVPR'22). Among them, DDMSNet was the most advanced algorithms with a parameter size of 229.454M. DesnowNet requires a mask dataset as input during training, which is unfortunately inaccessible in SnowKitti2012 and SnowCityScapes. Hence, we use the pre-trained weights of Snow100K for testing, denoted as \uuline{DesnowNet} in this paper. Besides, JSTASR does not provide training code except its pre-trained weights in SRRS. Therefore, we use it for all testing, denoted as \uline{JSTASR} in this paper.

\subsection{Comparison Results}

\subsubsection{Quantitative Results on Synthetic Dataset}
From the results of Table \ref{Table1}, our method achieves the state-of-the-art in most synthetic datasets. For instance, the PSNR of our method is 4.5$\%$ higher than the 2nd-best method on Snow100K dataset. On other synthetic datasets, our method also shows superior performances. The proposed method also has a lightweight design with only 2.181M parameters, which is 11.5$\%$ of the 2nd best method and 0.95$\%$ of DDMSNet. Meanwhile, the runtime of our method is only 0.042s, which is 1.21 times faster than the 2nd best method and 71.24 times faster than the lowest method. To conclude, these extensive experimental results show the superiority of our proposed method on synthetic datasets. 

\subsubsection{Qualitative Results on Synthetic Dataset}
Fig. \ref{Figure8} shows the visualized results of three synthetic snow removal datasets, including Snow100K, CSD and SnowCityScapes. DesnowNet and JSTASR fail to recover clean images and seriously damage the original snowy image. DDMSNet loses some color and clarity in details. HDCWNet preserves repairing traces in details of clean images. Compared with those methods, the images recovered by our proposed method are clean and intact with better detail preservation. The qualitative results indicate that our method could generate images with better visual qualities.

\subsubsection{Comparison on Large Snowflakes and Veiling Effect}
Large snowflakes and veiling effect are two challenge phenomenons in snow scenes. Snow100K provide large snowflakes dataset called Snow100K-L. SRRS and CSD datasets specifically provide snowy images with veiling effect. 

From the results of Snow100K-L, SRRS and CSD in Table \ref{Table1}, our algorithm achieves superior or comparable performance than other methods. Fig \ref{Figure9} shows that our algorithm effectively removes large snowflakes and veiling effect, while other methods fail to remove them completely. These demonstrate the effectiveness of our algorithm in removing large snowflakes and veiling effects.

\subsubsection{Comparison on Real-World Dataset}
Fig. \ref{Figure9} shows the effects of snow removal in real-world. The qualitative results show that other snow removal methods lose details and have repair effects, which do not work well in the real-world scenarios. In contrast, our algorithm has the best snow removal effect with almost no repairing traces in the details. It confirms the excellent generalization performance of our proposed algorithm in real-world.

\begin{table}[htbp]
\caption{Ablation results of different modules of Laplace-VQVAE.}
\label{Table2}
\centering
\renewcommand
\arraystretch{1.3}
\resizebox{0.485\textwidth}{!}{
    \begin{tabular}{ccccccc|ccc}
    \toprule
    \textbf{VAE} & \textbf{VQVAE} & \textbf{Conv} & \textbf{ConvK} & \textbf{CA} & \textbf{PA} & \textbf{SA} & \textbf{PSNR(↑)}                         & \textbf{SSIM(↑)} & \textbf{\begin{tabular}[c]{@{}c@{}} Parameters\end{tabular}} \\ \hline
    \checkmark   &   &     &     &    &   & \checkmark   & 29.551                                 & 0.934                                 & 9.63M                                                                    \\
                 & \checkmark              & \checkmark             &                &             &             &               & 29.694                                 & 0.937                                 & 92.22K                                                                   \\
                 & \checkmark              &               & \checkmark              &             &             &               & 29.997                                 & 0.94                                  & 156.73K                                                                  \\
                 & \checkmark              &               & \checkmark              & \checkmark           &             &               & 30.139                                 & 0.941                                 & 157.02K                                                                  \\
                 & \checkmark              &               & \checkmark              &             & \checkmark           &               & 30.194                                 & 0.94                                  & 157.2K                                           \\ \midrule
                 & \checkmark              &               & \checkmark              &             &             & \checkmark             & \textbf{30.286} & \textbf{0.942} & \textbf{157.32K}                                                         \\ \bottomrule
    \end{tabular}}
\end{table}

\subsection{Ablation Studies}
We explore the contributions of different modules and query number in our proposed method. Moreover, we conduct parameter sensitive analysis of loss function. Evaluation is performed on Snow100K dataset and our method is trained with 2\(\times\)10\(^5\) iterations.

\subsubsection{Contributions of Different Modules of Laplace-VQVAE}
\begin{figure*}[htbp]
    \centering
    \includegraphics[width=0.8\textwidth]{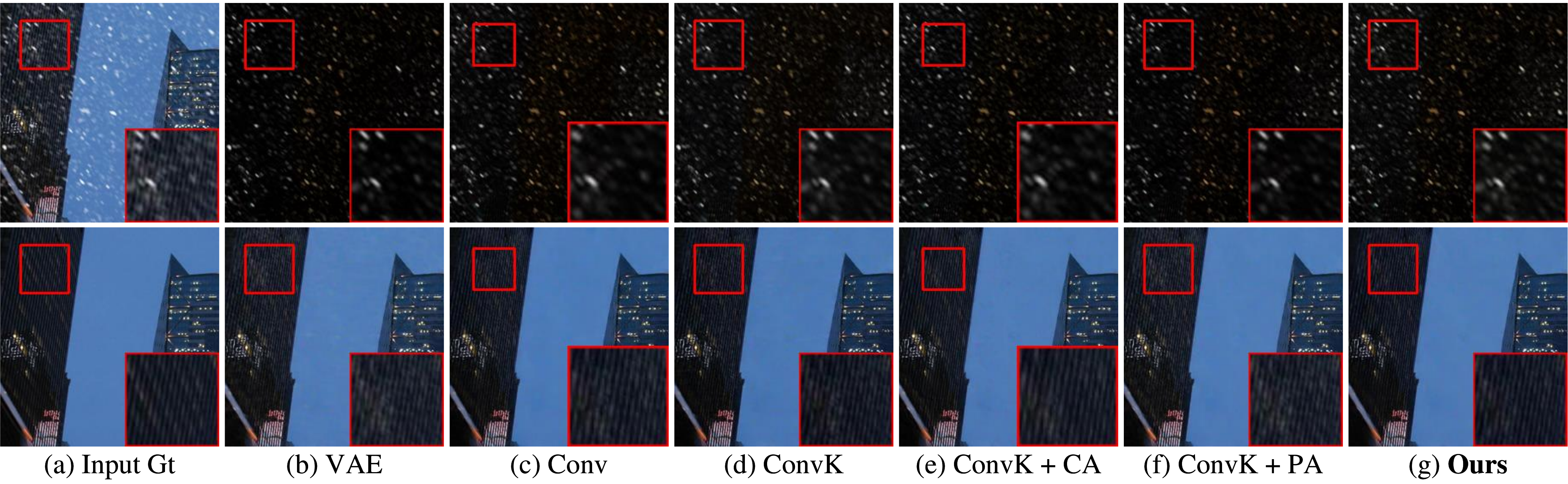}
    \caption{Qualitative results of different modules of Laplace-VQVAE.} 
    \label{Figure11}
\end{figure*}

We conduct ablation studies on Laplace-VQVAE, where \textbf{VAE} and \textbf{VQVAE} represent different backbones, \textbf{Conv} represents convolution with fixed kernel, \textbf{ConvK} represents convolution with different kernels, \textbf{CA}, \textbf{PA} and \textbf{SA} represent channel attention, pixel attention and spatial attention, respectively. To meet the VAE's requirement of fixed input size, we crop all input images to 256$\times$256 in our experiments. As shown in Table \ref{Table2}, our proposed Laplace-VQVAE has a parameter size of 157.32K, which is much smaller than VAE's 9.63M. Besides, SACM significantly improves the performance with a small increase of parameter size. Fig \ref{Figure11} indicates that our method learns more effective snow features compared with other modules.

\begin{figure}[]
    \includegraphics[width=0.45\textwidth]{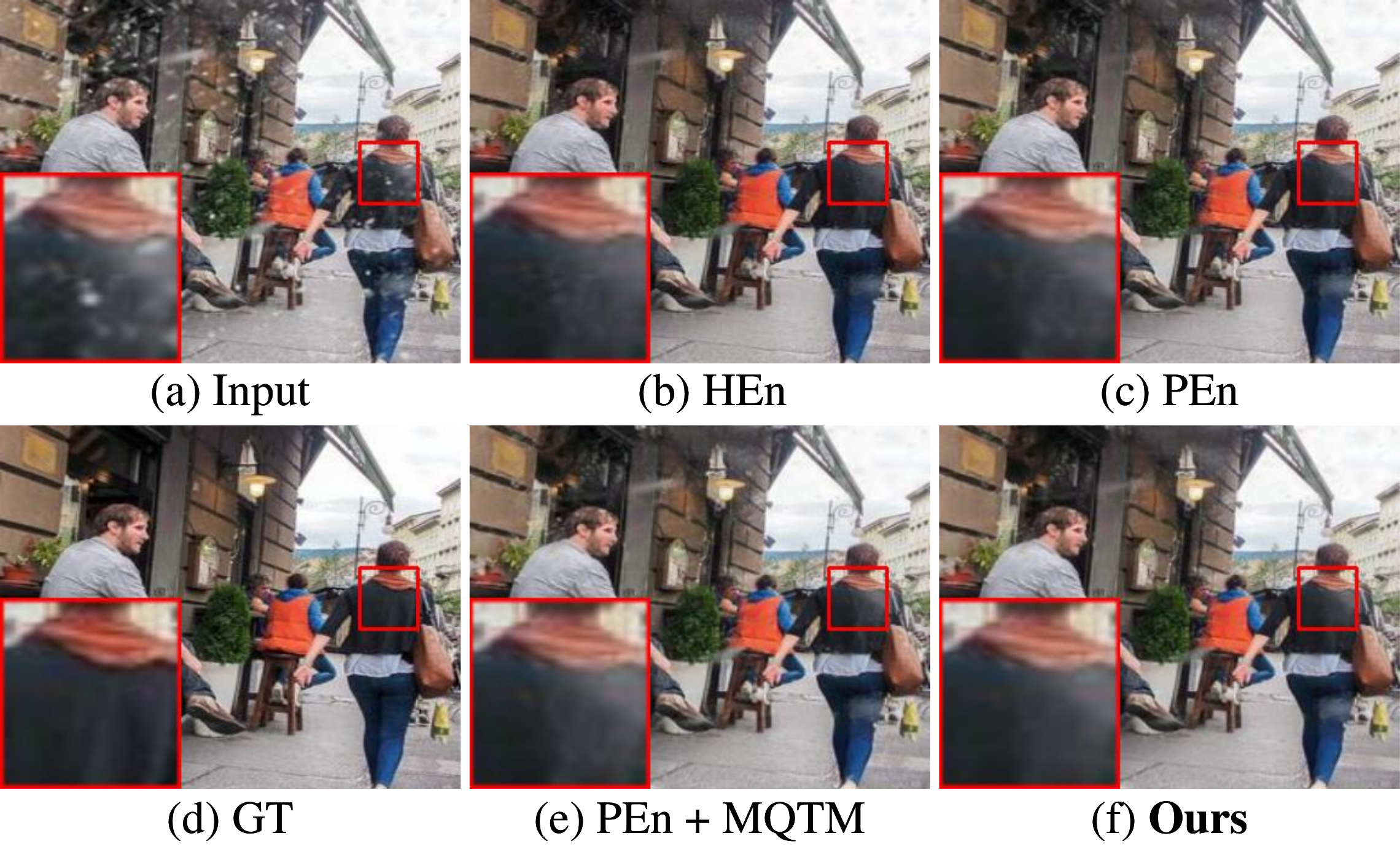}
    \centering
    \caption{Qualitative results with different settings of MQFormer. (b) H-En w/o Mask; (c) P-En w/o Mask; (e) P-En + MQTM w/o Mask; (f) Ours.}
    \label{Figure12}
\end{figure}

\begin{table}[]
\centering
\caption{Quantitative results with different settings of MQFormer.}
\label{Table3}
\centering
\renewcommand\arraystretch{1.5}
\scalebox{0.92}{
    \begin{tabular}{c|cccc}
    \toprule
    \textbf{Metric}     & \textbf{\makecell[c]{H-En \\w/o Mask}} & \textbf{\makecell[c]{P-En \\w/o Mask}} & \textbf{\makecell[c]{P-En + M C\\ w/o Mask}} & \textbf{\makecell[c]{P-En + M C\\w/ Mask (Ours)}} \\ \midrule
    \textbf{PSNR(↑)} & 27.771                                                   & 28.745                                                  & 28.371                                                       & \textbf{29.014}                                     \\ \hline
    \textbf{SSIM(↑)} & 0.895                                                    & 0.923                                                   & 0.919                                                        & \textbf{0.931}                                      \\
    \bottomrule
    \end{tabular}}
\end{table}

\subsubsection{Contributions of Different Modules of MQFormer}
Our proposed network consists of three parts: parallel encoders, lightweight modules and mask prior. Specifically, we design parallel encoders to learn features of different dimensions. MQTM and CACM are the main lightweight modules. Meanwhile, mask prior is also an important input feature of our model. We list the following configurations:

\textbf{H-En w/o Mask}: Hybrid encoder without the coarse mask as prior knowledge. The encoder has the same architecture as hybrid decoder.

\textbf{P-En w/o Mask}: Parallel encoders without the coarse mask as prior knowledge. They are transformer encoder and convolutional encoder.

\textbf{P-En + M C w/o Mask}: Parallel encoders without the coarse mask as prior knowledge. MQTM replaces the original vision transformer modules \cite{dosovitskiy2020image}. We add MLP and Batchnorm to convolutional modules as CACM.

\textbf{P-En + M C w/ Mask (Ours)}: Complete network architecture with the coarse mask as prior knowledge.

\begin{figure}[]
    \includegraphics[width=0.45\textwidth]{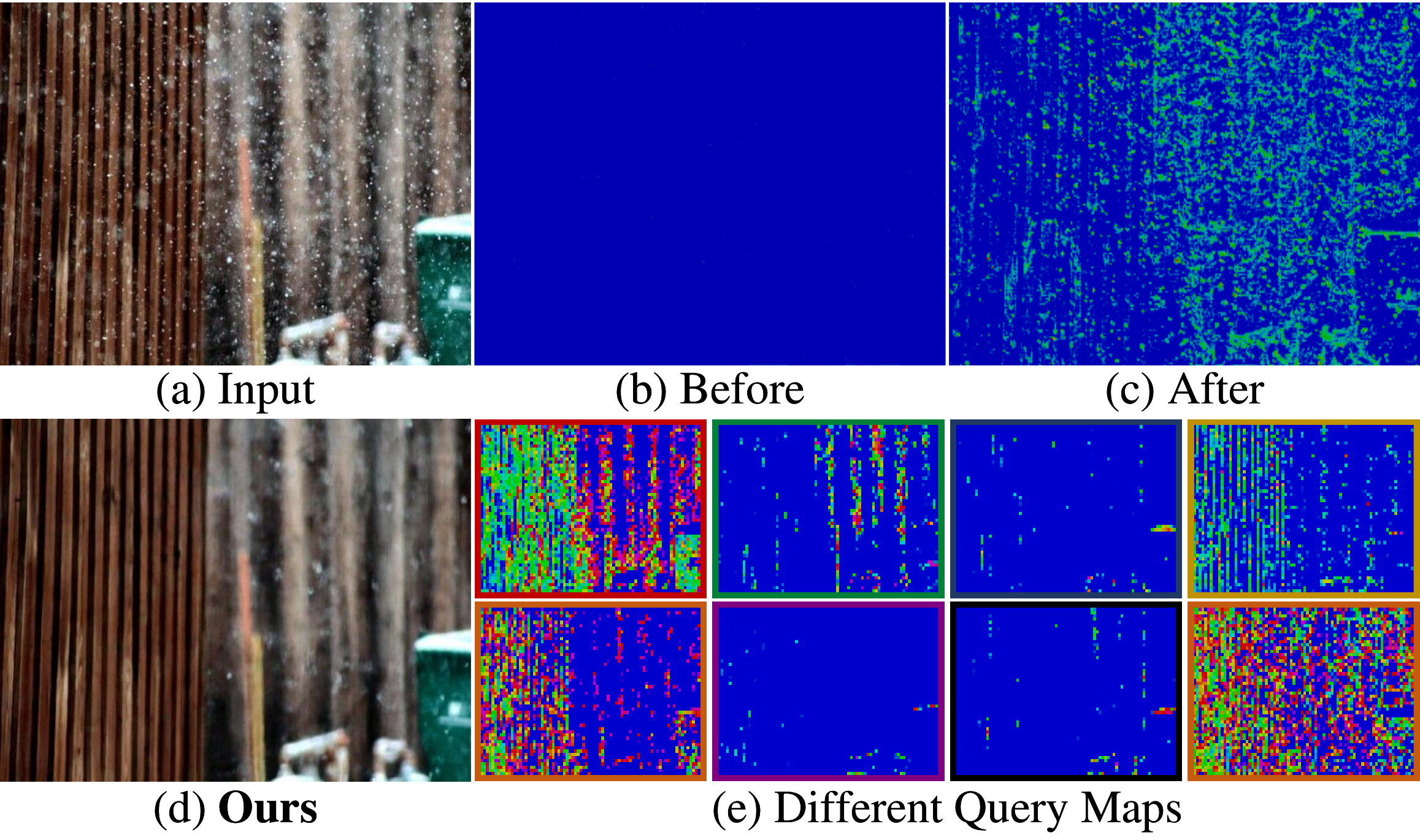}
    \centering
    \caption{Visualization of the feature maps.}
    \label{Figure13}
\end{figure}

From the results of Table \ref{Table3}, we can observe that each design has made a contribution to performance. Qualitative results are shown in Fig. \ref{Figure12}. The feature maps before and after MQTM are shown in Fig. \ref{Figure13}(b)(c). 

\begin{table}[]
\caption{Quantitative results with different query number $Q$.}
\label{Table4}
\centering
\renewcommand
\arraystretch{1.5}
\centering
\scalebox{0.92}{
    \begin{tabular}{c|cccc}
    \toprule
    \textbf{Metric} & \textbf{4 Q}      & \textbf{16 Q}      & \textbf{Original Q}     & \textbf{8 Q (Ours)}   \\ \midrule
    \textbf{PSNR(↑)} & 28.767  & 28.708 & 28.823  & \textbf{29.014} \\ \hline
    \textbf{SSIM(↑)} & 0.924 & 0.918 & 0.924 & \textbf{0.931}  \\ 
    \bottomrule
    \end{tabular}}
\end{table}

\subsubsection{Contributions of Different Query Numbers}
We set query number to 4, 16, original number and 8 (Ours) and compare the model performances under these settings. From Table \ref{Table4}, we observe a slight performance improvement with a smaller query number. We set it to 8 in our network. As shown in Fig. \ref{Figure13}(e), each query has learned concentrate feature maps, which confirms our main concept of treating Query as a specific number of mask filters to concentrate attention on snow areas.

\subsubsection{Parameter Sensitive Analysis of Loss Function}
We conduct ablation studies on the parameter sensitive analysis of loss function. We fix $\lambda_2=0.1$ and perform ablation studies on $\lambda_1$ with 0.025 and 0.1, respectively. Similarly, we fix $\lambda_1=0.05$ and perform ablation studies on $\lambda_2$ with 0.05 and 0.2, respectively. As shown in Table \ref{Table5}, our hyper-parameter setting has the best performance, indicating the effectiveness of our parameter setting.

\begin{table}[]
\caption{Quantitative results with different hyper-parameters of loss function.}
\label{Table5}
\centering
\renewcommand
\arraystretch{1.2}
\scalebox{1}{
    \begin{tabular}{cc|cc}
    \toprule
    \textbf{$\lambda_1$}                     & \textbf{$\lambda_2$}                    & \textbf{PSNR(↑)}                          & \textbf{SSIM(↑)}                         \\ \midrule
    0.025                           &                                & 29.275                                 & 0.927          \\
    0.1                             & \multirow{-2}{*}{\textbf{0.1}} & 29.248                                 & 0.926                                 \\ \hline
                                    & 0.05                           & 29.047                                 & 0.924                                 \\
    \multirow{-2}{*}{\textbf{0.05}} & 0.2                            & 29.242                                 & 0.927         \\ \midrule
    \textbf{0.05}                   & \textbf{0.1}                   & \textbf{29.279} & \textbf{0.927} \\ \bottomrule
    \end{tabular}}
\end{table}

\section{Conclusion}
Snow removal task is critical in computer vision under snowy weather due to the distinct visual appearance and feature distribution of snow. In this paper, we propose a lightweight but high-efficient snow removal network called LMQFormer. It consists of a Laplace-VQVAE sub-network to reduce redundant information entropy with the coarse mask and an MQFormer sub-network to recover clean images without repairing traces. We propose a mask query transformer module to concentrate attention to snow areas. Experimental results show that our proposed LMQFormer achieves superior performances on existing snow removal datasets, whilst its parameter size is significantly reduced compared with its peers.

\normalem
\bibliographystyle{IEEEtran}
\bibliography{arxiv}

\end{document}